\documentclass{Interspeech2024}
\usepackage{multirow}
\usepackage[T1]{fontenc}
\usepackage{booktabs}
\usepackage[newcommands]{ragged2e}
\usepackage{graphicx,caption} 
\usepackage{svg}
\usepackage{cite}
\usepackage{hyperref}

\interspeechcameraready

\title{Rapport-Driven Virtual Agent: Rapport Building Dialogue Strategy for Improving User Experience at First Meeting}

\name[affiliation={1,2}]{Muhammad Yeza}{Baihaqi}
\name[affiliation={2}]{Angel García}{Contreras}
\name[affiliation={1,2}]{Seiya}{Kawano}
\name[affiliation={1,2}]{Koichiro}{Yoshino}

\address{
  $^1$Nara Institute of Science and Technology, Japan\\
  $^2$Guardian Robot Project, RIKEN, Japan 
}

\email{muhammad\_yeza.baihaqi.lx2@naist.ac.jp, angel.garciacontreras@riken.jp, seiya.kawano@riken.jp, koichiro.yoshino@riken.jp}


\begin{document}

%maketitle
\maketitle

\begin{abstract}
Rapport is known as a conversational aspect focusing on relationship building, which influences outcomes in collaborative tasks. This study aims to establish human-agent rapport through small talk by using a rapport-building strategy. We implemented this strategy for the virtual agents based on dialogue strategies by prompting a large language model (LLM). In particular, we utilized two dialogue strategies—predefined sequence and free-form—to guide the dialogue generation framework. We conducted analyses based on human evaluations, examining correlations between total turn, utterance characters, rapport score, and user experience variables: naturalness, satisfaction, interest, engagement, and usability. We investigated correlations between rapport score and naturalness, satisfaction, engagement, and conversation flow. Our experimental results also indicated that using free-form to prompt the rapport-building strategy performed the best in subjective scores. 
\end{abstract}

\section{Introduction}
Human-agent interaction (HAI) is a burgeoning field, exploring the dynamics between individuals and artificial agents. Types of agents vary widely, encompassing virtual agents and physical agents such as robots \cite{1}. Task-oriented HAI systems have been widely researched recently; however, only focusing on abilities directly related to task outcomes is insufficient \cite{2}; it can lead to a less engaging user experience, in other words. Conversely, emphasizing relational aspects such as rapport has the potential to notably improve both user experience and task outcomes.

Rapport is defined as a cordial and easeful connection characterized by mutual understanding, acceptance, and sympathetic compatibility between or among individuals \cite{3}. In collaborative tasks, rapport is key for effective communication, trust, and a positive atmosphere. It ensures sustained engagement, interest, and satisfaction among team members. Existing research emphasizes the critical importance of establishing rapport to enhance task outcomes in many applications, including tutoring \cite{4}, food delivery \cite{5}, healthcare \cite{6}, and negotiation \cite{nego}. 

One effective way identified for cultivating rapport is using small talk for ice-breaking\cite{7, icebreaker2}. To establish rapport between humans and agents, the small talk interaction has to be engaging \cite{8, 27}. However, we do not have a concrete way to introduce such rapport-building in dialogue. In this research, we focus on small talk at first-time meetings, a situation that requires rapport building to establish a relationship for outcomes of descendant dialogue. 

In this study, we introduced a novel rapport-building dialogue strategy on small talk, for enhancing relationships during the ice-breaking. Our frameworks were implemented as guides for a large language model (LLM)-based small talk system to realize cooperative rapport building. Our study employed both predefined sequence and free-form dialogue strategies, infusing the virtual agent's small talk with rapport-building utterances. The predefined sequence strategy introduced structure and predictability, guiding conversations along predetermined paths to ensure controlled interactions. In contrast, the free-form strategy injected spontaneity and adaptability, capturing the nuances of natural language for a more dynamic interaction. 

We systematically compared these strategies through human evaluation using a virtual agent platform, ERICA. The experimental results showed that our rapport-building strategies significantly improved some metrics related to user experience variables. We also explored possible factors influencing rapport by analyzing the correlation between rapport score, total turns, total utterance characters, and user experience variables.

\section{Related work}
Small talk is often described as a casual conversation that is deemed trivial; however, it serves a crucial purpose in cultivating social bonds among people \cite{9}. Within the realm of HAI, the integration of small talk functionalities has been a recurrent focus. Numerous studies have illuminated its advantageous outcomes, demonstrating its efficacy in enhancing engagement levels, bolstering dialog satisfaction, cultivating interest in ongoing conversations, and fortifying the establishment of rapport between humans and agents \cite{10,11}. 

However, prevalent research in this domain has primarily implemented small talk within the context of ongoing tasks or activities \cite{12}. In certain practical applications, constraints dictate that small talk occurs exclusively before the initiation of a task, especially during the first meeting dialogues. Furthermore, the existing investigation has primarily confined the scope of small talk to conventional topics such as weather \cite{2}. While adequate for filling intervals between tasks, recent scholarly attention has pivoted towards augmenting small talk capabilities to foster more intricate and meaningful connections between agents and humans \cite{2, 28}. Prior research limited small talk to Q\&A formats, causing artificial flow with a single sequence; hence, the agent needed varied strategies for small talk across different complexities \cite{2,26}. Two notable studies diverged from the Q\&A sequence; however, their approach involved the control of the agent through the Wizard-of-Oz (WoZ) \cite{5, 13}. Lastly, the recent findings in \cite{11} have highlighted a limitation wherein individuals engaging with agents were unable to pose their questions during small talk sessions.

\section{Proposed method}

\subsection{Rapport-building dialogue strategy}
Our research introduced a rapport-building dialogue strategy by integrating rapport-building utterances into the small talk with a virtual agent which was gathered from various existing studies of rapport-building \cite{10,20,21,22,23,24}. These carefully curated utterances, derived from prior research demonstrating remarkable success in fostering rapport across diverse contextual domains. By adapting these varied utterances, our goal was to realize the benefits associated with each utterance and enrich the communication, ultimately leading to higher human-agent rapport. These utterances are:

\begin{itemize}
    \item Participation appreciation \cite{21}: Acknowledging and valuing individuals' contributions.
    \item Praise expression \cite{24}: Fostering a supportive and affirming atmosphere, and encouraging continued active engagement.
    \item Self-disclosure \cite{23}: Creating a sense of openness.
    \item Knowledge sharing \cite{20}: Demonstrating expertise, establishing credibility, and fostering a sense of trust.
    \item Empathetic response \cite{20,24}: Demonstrating understanding, validation, and genuine interest.
    \item Storytelling \cite{10}: Building trust between human and agent by sharing personal experience.
    \item Recommendation giving \cite{20}: Demonstrating a desire to addressing specific needs.
    \item Positive encouragement \cite{24}: Creating a supportive and encouraging environment.
    \item Joke sharing \cite{20,24}: Creating an enjoyable atmosphere.
    \item Name usage \cite{22}: Personalizing the interaction and creating a sense of recognition.
\end{itemize}
In addition, our agent delivered both close-ended and open-ended questions at intermittent intervals within the dialogue to sustain its continuity. Close-ended questions elicited concise responses, ensuring conversational fluidity, while open-ended questions encouraged detailed reflections, enriching the depth and complexity of the dialogic interaction \cite{25}.

The rapport-building dialogue strategy was embedded into the agent as different strategy implementations, free-form and predefined dialogue strategies. In the free-form strategy, the virtual agent gained the advantage of fostering a more natural and dynamic conversation, allowing users to express themselves authentically. This approach offered flexibility and adaptability, enhancing user engagement by responding to unique cues. However, drawbacks included potential inconsistency and missed opportunities for strategic rapport-building. On the other hand, predefined elicitation ensured consistency and goal alignment but led to a more rigid and less personalized dialogue. In particular, we implemented them as one-by-one prompts for each utterance or a whole prompt for the whole dialogue. 

We also defined Rapport Score (RS) based on an existing study \cite{18}. The score is calculated from human evaluation scores; an average of scores for 7-point Likert scale questionnaires defined in Table~\ref{tab:x}.

\begin{table}[t!]
  \caption{Questionnaire to measure rapport}
  \label{tab:x}
  \renewcommand{\arraystretch}{1} % Adjust the vertical padding
  \centering
  \begin{tabular}{ll}
    \toprule
    \textbf{\footnotesize ID}                  
    & \textbf{\footnotesize Questions}             
    \\
    \midrule
    \footnotesize R1 & \footnotesize I think about my relationship with this virtual agent.\\
    \footnotesize R2 & \footnotesize I enjoyed interacting with this virtual agent. \\
    \footnotesize R3 & \footnotesize This virtual agent is very relevant to me. \\
    \footnotesize R4 & \footnotesize I felt comfortable interacting with this virtual agent. \\
    \footnotesize R5 & \footnotesize I feel a bond between this virtual agent and myself. \\
    \footnotesize R6 & \footnotesize I really care about this virtual agent. \\
    \footnotesize R7 & \footnotesize This virtual agent has a personal interest in me. \\
    \bottomrule
  \end{tabular}
\end{table}

\subsection{Multimodal virtual agents}

\subsubsection{Virtual agent architecture}
In this study, we employed a female virtual agent, ERICA, who interfaced with seven modules \cite{19}. These modules encompassed Japanese automatic speech recognition (ASR) for processing human speech into text, a language model for contextual understanding and coherent response generation, Japanese text-to-speech (TTS) for converting textual output to speech, LipSync for synchronized lip movement with speech, gesture and face expression control for facial and gestural regulation, and a conversation database for storing dialogue log. 

The agent maintained a smiling expression throughout interactions and exhibited a proper sitting posture. Notably, the virtual agent incorporated a bowing gesture, a Japanese-style greeting, both at the beginning and end of conversations.

In this research, we adopted two distinct dialogue strategies: free-form interaction and a predefined scenario. For the agent developed through a free-form strategy, we harnessed the power of LLMs. Employing a chat completion function, the model was exclusively prompted at the initiation of the conversation. Subsequently, the model autonomously crafted responses, consistently referencing the initial command throughout the conversation facilitating a more dynamic and adaptable dialogue for building rapport. 

In constructing our agent based on a predefined scenario, we chose the completion type of GPT-3.5-Turbo, GPT-3.5-Turbo-Instruct. This model prompted with a completion function at each dialogue turn, was preferred for its adeptness in handling focused single-input tasks over chat completion models. Unlike chat completion, which possibly introduced contextual ambiguity, the completion model prioritized simplicity for predefined sequences. To maintain contextual coherence, utterances from preceding turns were integrated into commands during subsequent turns. 

\subsubsection{Virtual agent types}
The four agents were built using the architecture explained in subsection 3.2.1, each distinguished by its dialogue strategy and dialogue turn limit. The agent with the term ``rapport'' in its name indicated the inclusion of a rapport-building dialogue strategy. The rapport-building strategy was included in the LLM by prompting\footnote{The link to the prompt is available on \url{https://github.com/yezato11/INTERSPEECH_Media}}. 

First, the Limit Free Rapport Agent was set without a turn limit. This intentional decision allowed for a more comprehensive exploration of the correlation between the total utterances characters uttered by the agent and participant, total turn, and RS with the user experience (UX) variables. These UX variables are further elaborated upon in subsection 4.4.

Free Rapport Agent also adopted a free-form dialogue strategy. Nevertheless, a distinctive characteristic of Free Rapport Agent lay in its implementation of a fixed turn limit, capped at 20 turns. Once, it reached 20 turns, the agent finished the small talk. This intentional limitation set Free Rapport Agent apart and positioned it for subsequent comparative analyses. This agent employed a rapport-building dialogue strategy, however the order of the utterances was randomized, leaving the approach's performance uncertain over multiple dialogues. 

The Predefined Rapport Agent incorporated a rapport-building strategy using a predefined scenario that concluded after 20 turns. This agent utilized all the strategies in section 3.1, following the order presented therein. This agent could effectively use a rapport-building dialogue strategy when the predefined scenario matched the user, but sticking to the scenario posed a risk of missing personalized conversation opportunities.

Q\&A Agent represented the traditional Q\&A agent. In contrast, this agent exclusively posed questions and provided simple acknowledgment based on the context of the conversation. 

\section{Experimental settings}

\subsection{Participants}
We conducted dialogue experiments with participants. A total of 20 participants, consisting of 11 males and 9 females, were recruited for the experiment. All participants were graduate students and native speakers of Japanese. None of the participants had prior experience with virtual agent experiments. The average age of the participants was 24.75 years (\textit{SD} = 2.7). In addition, before the study,  participants willingly granted informed consent for the inclusion of their data in the research.

\subsection{Small talk topics}
Selecting suitable small talk topics was crucial for fostering positive interactions. Opting for light yet intimate subjects was a key consideration. In our research, we chose dream travel destinations for their blend of personal passion and universal appeal, making them ideal for fostering engaging conversations. Moreover, this topic served as a versatile conversation. It smoothly led to discussions about local cuisine, iconic landmarks, and compelling reasons for visiting the destination.

\subsection{Experimental procedure}
In our research, we used a counterbalancing approach, where each participant interacted with and rated all virtual agents. To prevent any potential bias from the order of the agent, we randomized the sequence. 

\subsection{Questionnaire and analyses}
This experimental study rigorously scrutinized the agents through a multifaceted analysis facilitated by a structured questionnaire, employing a 7-point Likert scale for human evaluation. The questionnaire systematically gauged various UX variables, encompassing naturalness (N1), satisfaction (S2), interest (I3), engagement (E4, E5), and usability (U6, U7) as shown in Table~\ref{tab:1}. In this research, three analyses were conducted. 

\begin{enumerate}
\item \textbf{Correlation analysis}: We analyzed the correlation of RS, total turn, and the total utterance characters by the agent and participant with UX variables. Limit Free Rapport Agent was utilized for this analysis. In addition, participants were allowed to ask questions. 

\item \textbf{Strategy comparison}: We compared the Free Rapport Agent with the Predefined Rapport Agent to investigate the better dialogue strategy in the context of UX variables and RS. To uphold fairness during the evaluation, participants were instructed not to pose questions to agents. This decision stemmed from the consideration that the Predefined Rapport Agent operated within a predefined scenario, and allowing participants to ask questions could potentially create an unfair disadvantage. 

\item \textbf{Effect of using rapport}: We compared the best Rapport Agents from strategy comparison analysis with the Q\&A agent to investigate the effect of using a rapport-building dialogue strategy. 
\end{enumerate}

\begin{table}[b!]
  \caption{Questionnaire to measure UX variables}
  \label{tab:1}
  \renewcommand{\arraystretch}{1} % Adjust the vertical padding
  \centering
  \begin{tabular}{>{\centering\arraybackslash}m{0.03\linewidth}>{\justifying\raggedright\arraybackslash}p{0.85\linewidth}}
    \toprule
    \textbf{\footnotesize ID}                  
    & \textbf{\footnotesize Questions}             
    \\
    \midrule
    \footnotesize N1 & \footnotesize Conversations with virtual agents felt natural.\\
    \footnotesize S2 & \footnotesize I am satisfied with my conversation with the virtual agent.\\
    \footnotesize I3 & \footnotesize The conversation with the virtual agent was interesting.\\
    \footnotesize E4 & \footnotesize The conversation with the virtual agent was engaging.\\
    \footnotesize E5  & \footnotesize I would like to continue the dialogue with the virtual agent next time.\\
    \footnotesize U6 & \footnotesize Conversations with virtual agents were easy to understand. \\
    \footnotesize U7 & \footnotesize Conversations with virtual agents maintained a logical flow.\\
    \bottomrule
  \end{tabular}
\end{table} 

\section{Results}
\subsection{Correlation analysis}
From our experimental investigation, we observed a notable pattern among participants as they engaged in small talk, with an average of 27.35 turns (\textit{SD} = 7.82). The communication output, encompassing both agent and human utterances, revealed an average of 1734 total utterance characters (\textit{SD} = 551.65). 

As shown in Table~\ref{tab:2}, our works revealed that RS exhibited significant correlations with UX variables influencing the conversational experience. Employing the two-tail Spearman correlation test, we found a moderate correlation between RS and the naturalness of the dialogue ($\rho$ = 0.525, \textit{p} < 0.05). Moreover, RS was found to have a strong correlation with satisfaction ($\rho$ = 0.662, \textit{p} < 0.01) and engagement, showing robust associations with engagement in E4 and E5 ($\rho$ = 0.706, \textit{p} < 0.01, $\rho$ = 0.661, \textit{p} < 0.01). Additionally, a moderate correlation was identified between RS and the usability of the dialogue, specifically concerning the logical flow ($\rho$ = 0.512, \textit{p} < 0.05). Notably, this study revealed that RS did not exhibit any correlation with the interest in the conversation and the perceived easiness of understanding the dialogue. Additionally, the UX variables and RS showed no correlation with the total turns and characters uttered by both the participant and the virtual agent.

\begin{table*}[t!]
    \footnotesize
  \caption{Correlation analysis of RS, total turn of the small talk, and the total utterances characters with UX variables}
  \label{tab:2}
  \renewcommand{\arraystretch}{1} % Adjust the vertical padding
  \centering
  \begin{tabular}{p{0.225\linewidth}llllllll}
    \toprule
     & \textbf{N1} & \textbf{S2} & \textbf{I3} & \textbf{E4 }& \textbf{E5} & \textbf{U6} & \textbf{U7} & \textbf{RS} \\
    \midrule
    \textbf{RS} & 0.525*& 0.662** & 0.426 & 0.706** & 0.661** & 0.271 & 0.512* & 1\\
    \textbf{Total Turn} & 0.200 & -0.210 & 0.290 & -0.144 & 0.058 & -0.114 & -0.287 & 0.063\\
    \textbf{Total Utterances Characters} & 0.127& -0.017 & 0.284 & -0.164 & -0.023 & -0.217 & -0.265 & 0.032\\
    \bottomrule
    \multicolumn{9}{r}{*\textit{p} < 0.05, **\textit{p} < 0.01}
    \end{tabular}
\end{table*}

\subsection{Strategy comparison}
\begin{figure}[t!]
  \centering
  \includegraphics[width=\linewidth]{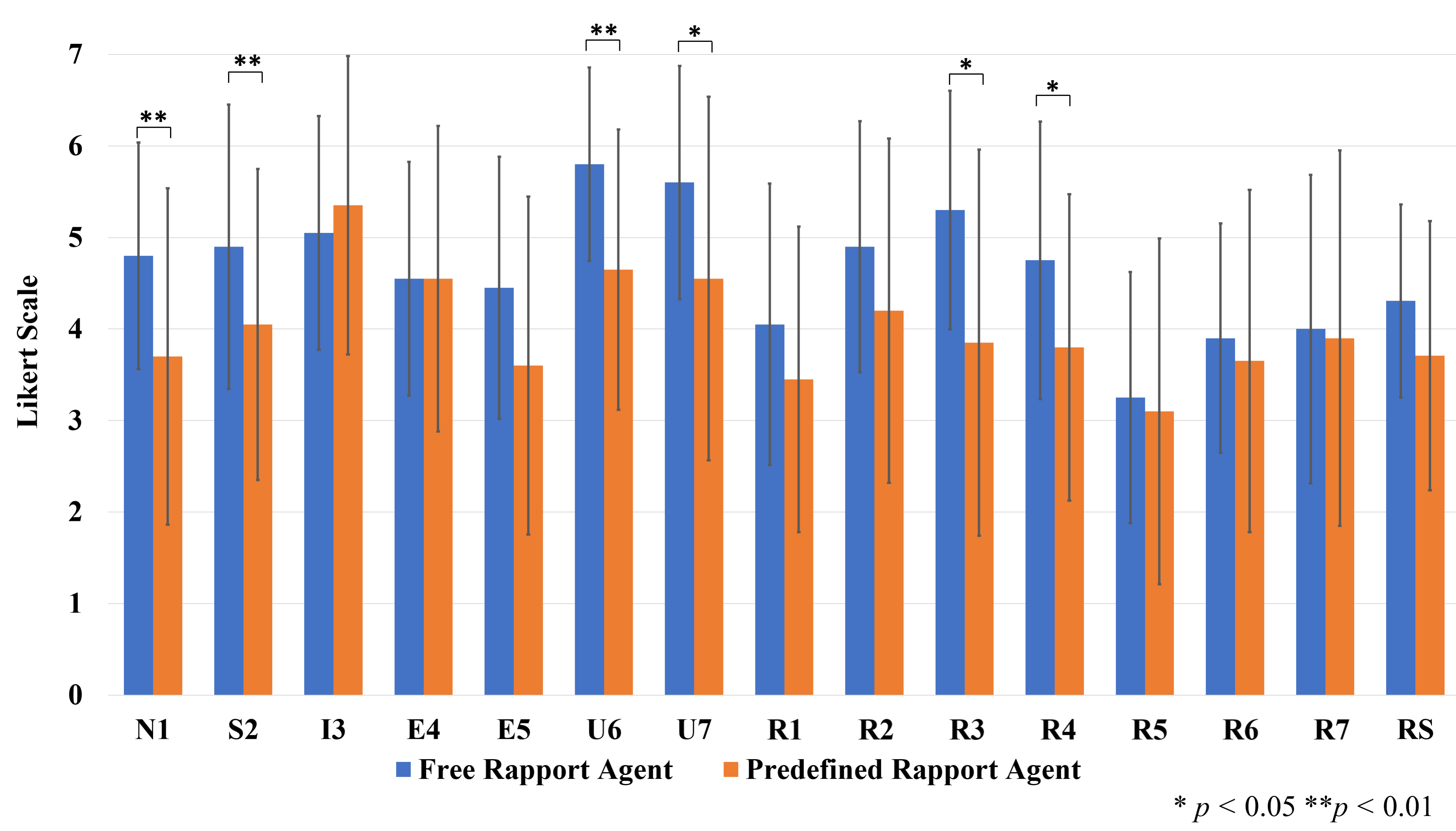}
  \caption{Comparative analysis between Free Rapport Agent and Predefined Rapport Agent.}
  \label{fig:2}
\end{figure}

The experimental data unveiled compelling findings regarding the performance of the Free Rapport Agent in comparison to the Predefined Rapport Agent, as shown in Figure~\ref{fig:2}. Our findings were particularly noteworthy, as they revealed that the Free Rapport Agent consistently outperformed the Predefined Rapport Agent across UX variables with statistical significance determined by the two-tailed Wilcoxon test. Specifically, in N1, S2, U6, and U7, the Free Rapport Agent garnered higher ratings, as evidenced by \textit{z}-scores of -2.614, -2.678, -2.615, and -2.430, respectively, with N1, S2, and U6 associated \textit{p}-values falling less than 0.01 and U7 associated \textit{p}-values falling less than 0.05. While no discernible differences emerged in RS,  however, in R3 and R4, the Free Rapport Agent exhibited a significant difference, shown by \textit{z}-scores of -2.320 and -2.096, respectively, with \textit{p}-values less than 0.05.

\subsection{Effect of using rapport-building dialogue strategy}
\begin{figure}[t!]
  \centering
  \includegraphics[width=\linewidth]{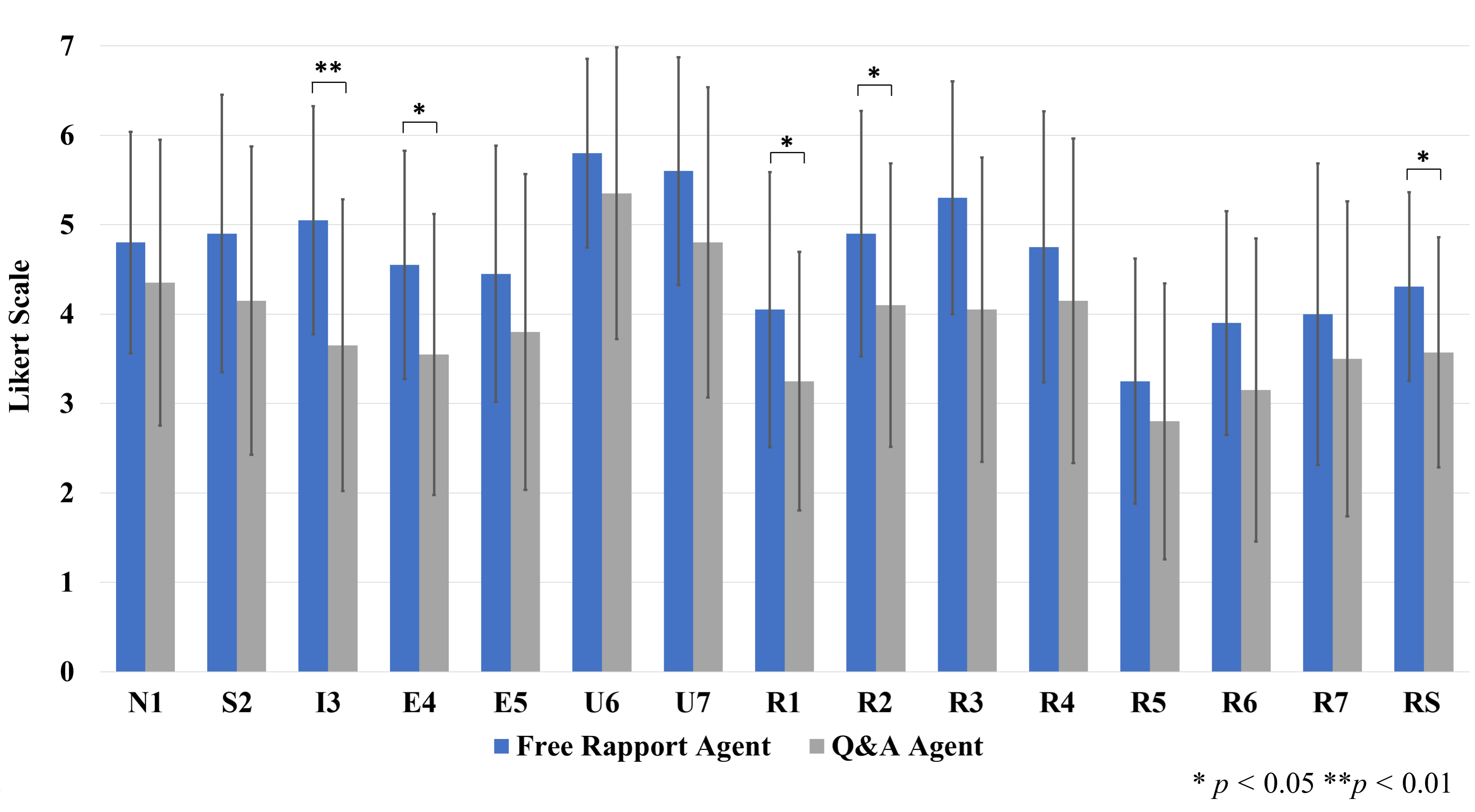}
  \caption{Comparative analysis between Free Rapport Agent and Q\&A Agent.}
  \label{fig:3}
\end{figure}

Figure~\ref{fig:3} highlighted a notable disparity in the performance between Free Rapport Agent and Q\&A Agent across all UX variables. The statistical analysis, utilizing a two-tail Wilcoxon test in I3 (\textit{z} = -2.930, \textit{p} < 0.01) and E4 (\textit{z} = -2.196, \textit{p} < 0.05), indicated a significant difference, with Free Rapport Agent consistently exhibiting higher ratings. Furthermore, in the evaluation of rapport, it demonstrated superior ratings in R1 (\textit{z} = -2.215, \textit{p} < 0.05), R2 (\textit{z} = -2.200, \textit{p} < 0.01), and RS (\textit{z} = -2.114, \textit{p} < 0.05).

\section{Discussion}
Our study investigated significant correlations between UX variables; the RS correlated with some metrics such as the naturalness of the agent, satisfaction with the conversation, interest and engagement with the agent, and logical flow of the conversation. The absence of correlation between RS and UX variables with total speaking turns and utterance characters indicated that the rapport building was not solely dependent on quantitative aspects. Interestingly, high RS was reported by participants who gained new insights about their dream travel destination, in our qualitative analysis. This highlights that qualitative aspects also played a crucial role in shaping rapport. 

In the comparison of dialogue strategies, between the Free Rapport Agent and the Predefined Rapport Agent, the Free Rapport Agent was more favorable due to its adaptivity. Predefined sequences could make the conversation less fluid. For example, if a user expressed interest in visiting Europa Park, the Predefined Rapport Agent might not transition to discuss into specific aspect of Europa Park, as it had to adhere to a predefined sequence. In contrast, Free Rapport Agent's flexibility allowed for a dynamic exploration of user interests \footnote{The link to the dialogue sample is available on \url{https://github.com/yezato11/INTERSPEECH_Media}}. The limitations of predefined sequences may negatively impact naturalness, satisfaction, and usability. This reason also led Free Rapport Agent to be more relevant and more comfortable to interact with.

Moving beyond the Q\&A Agent, the Free Rapport Agent, incorporating rapport-building utterances, provided more diverse responses. With information sharing, recommendations, storytelling, and jokes, Free Rapport Agent made the interaction interesting and enjoyable. Participants noted gaining new information and even laughed at the agent's jokes. By using empathetic responses, positive reinforcement, addressing participants by their names, and praising their dream travel destinations, Free Rapport Agent enhanced engagement and prompted participants to reflect on their relationship with the agent. A participant openly expressed contentment when reassured about overcoming language barriers to visit London. Additionally, several participants acknowledged a heightened sense of closeness when the virtual agent addressed them by their names.         
 
\section{Conclusions}
In this study, we introduced the rapport-building dialogue strategy on building of virtual agent that conducted ice-breaking small talk in first meeting dialogues. Our experimental results indicated the positive influence of rapport-building utterances, evidenced by higher questionnaire ratings. Compared to Predefined Rapport Agent, Free Rapport Agent showed superior ratings, with significant differences in naturalness, satisfaction, usability, and rapport aspects. Additionally, when compared to Q\&A Agent, Free Rapport Agent outperformed, particularly in interest, engagement, and rapport. Notably, the rapport score correlated with some UX variables but exhibited no correlation with the total turn and utterance characters.

While some reported feeling happy and relieved, the potential of both positive reinforcement and empathetic responses was not fully explored. Participants noted that the impact was limited by the neutral tone of the agent. Future research could replicate the study using emotional TTS and explore non-verbal cues such as nodding to improve human-agent interaction.

\section{Acknowledgements}
This research was supported by JSPS KAKENHI Grant Number 22H04873, 22K17958, and 23K24910.

\bibliographystyle{IEEEtran}
\bibliography{template}

\end{document}